\documentclass[10pt,twocolumn,letterpaper]{article}

\usepackage[pagenumbers]{cvpr} 

\usepackage{graphicx}
\usepackage{amsmath}
\usepackage{amssymb}
\usepackage{multirow, booktabs}
\usepackage{makecell}
\usepackage{graphicx}
\usepackage{pifont}
\newcommand{\cmark}{\ding{51}}%
\usepackage{xcolor}
\usepackage{tablefootnote}

\usepackage[pagebackref,breaklinks,colorlinks]{hyperref}

\usepackage[capitalize]{cleveref}
\crefname{section}{Sec.}{Secs.}
\Crefname{section}{Section}{Sections}
\Crefname{table}{Table}{Tables}
\crefname{table}{Tab.}{Tabs.}

\def\confName{CVPR}
\def\confYear{2022}

\begin{document}

\title{ Nexar Dashcam Collision Prediction Dataset and Challenge }

\author{Daniel C. Moura\\
Nexar Inc.\\
{\tt\small daniel.moura@getnexar.com}
\and 
Shizhan Zhu\\
Nexar Inc.\\
{\tt\small shizhan.zhu@getnexar.com}
\and
Orly Zvitia\\
Nexar Inc.\\
{\tt\small orly.zvitia@getnexar.com}
}
\maketitle
\begin{abstract}
This paper presents the Nexar Dashcam Collision Prediction Dataset and Challenge, designed to support research in traffic event analysis, collision prediction, and autonomous vehicle safety. The dataset consists of 1,500 annotated video clips, each approximately 40 seconds long, capturing a diverse range of real-world traffic scenarios. Videos are labeled with event type (collision/near-collision vs. normal driving), environmental conditions (lighting conditions and weather), and scene type (urban, rural, highway, etc.). For collision and near-collision cases, additional temporal labels are provided, including the precise moment of the event and the alert time, marking when the collision first becomes predictable.

To advance research on accident prediction, we introduce the Nexar Dashcam Collision Prediction Challenge, a public competition on top of this dataset. Participants are tasked with developing machine learning models that predict the likelihood of an imminent collision, given an input video. Model performance is evaluated using the average precision (AP) computed across multiple intervals before the accident (i.e. 500 ms, 1000 ms, and 1500 ms prior to the event), emphasizing the importance of early and reliable predictions. 

The dataset is released under an open license with restrictions on unethical use, ensuring responsible research and innovation.

\end{abstract}

\section{Introduction}
\label{sec:intro}

\noindent Traffic accidents cause significant loss of life and financial damage worldwide each year.
Preventing various types of traffic accidents, such as vehicle collisions, vehicle-pedestrian collisions, and single-vehicle loss of control, can save lives and protect assets.
Research in autonomous driving and ADAS (Advanced Driver Assistance Systems)\cite{Geiger2012CVPR, cordts2015cityscapes} shows promising potential for improving driving safety.
This progress is driven by the widespread adoption of dashcams, which provide a field of view comparable to or even more comprehensive than that of a human driver.
Early anticipation of imminent traffic accidents\cite{chan2017anticipating} adds significant value to autonomous driving systems.
These systems can continuously analyze dashcam footage in real-time, aiming to make timely and confident predictions about potential accidents affecting the vehicle or occuring nearby. 

\begin{table}[ht]
\caption{Summary of the existing datasets and their properties: total number of video clips and frames, availability of attributes annotations, resolution, S/R (synthetic or real), and if open-sourced (``A'' indicates academic use only). MMAU~\cite{fang2022cognitive, fang2024abductive} also provides re-annotations the CCD~\cite{bao2020uncertainty}, DoTA~\cite{yao2020and, yao2022dota} and A3D~\cite{yao2019unsupervised} datasets.
 }

\resizebox{0.48\textwidth}{!}{\begin{tabular}{lcccccc}
\Xhline{4\arrayrulewidth}
      & \#Clips & \#Frames & Attr. & Res. & S/R & \begin{tabular}[c]{@{}c@{}}Open\\ source\end{tabular} \\ \hline
      DAD~\cite{chan2017anticipating}   &   1,750      &    175K      &     -         & \textbf{720}$\times$\textbf{1280}   &    \textbf{R}     &  A   \\  
      SA~\cite{zeng2017agent}   &   1,733      &    173K      &      -        & \textbf{720}$\times$\textbf{1280} &  \textbf{R}      &  -   \\  
      EpicFail~\cite{zeng2017agent}   &  3,000       &    $\approx$300K      &    - & -           &  \textbf{R}       &  A   \\  
      VIENA$^2$~\cite{aliakbarian2018viena}   &  \textbf{15K}       &    \textbf{2.25M}      &        - &   \textbf{1280}$\times$\textbf{1920}      &   S      &  {\color{blue} \cmark}    \\  
      NIDB~\cite{kataoka2018drive, suzuki2018anticipating} & 6,244 & \textbf{1.3M} &  - & -   & \textbf{R} & - \\ 
      A3D~\cite{yao2019unsupervised}   &  1,500       &  208K         &     {\color{blue} \cmark}\tablefootnote{Re-annotated by MMAU~\cite{fang2024abductive}.}   &    \textbf{720}$\times$\textbf{1280}    &   \textbf{R}      &  -\tablefootnote{The YouTube videos following the URLs are no longer available.}    \\  
      CTA~\cite{you2020traffic}   &  1,935       & 853K         &    - &     \textbf{720}$\times$\textbf{1280}        &    \textbf{R}     &   {\color{blue} \cmark}     \\
      CCD~\cite{bao2020uncertainty}   &  4,500      &   225K &      {\color{blue} \cmark}     &  \textbf{720}$\times$\textbf{1280}       &   \textbf{R}      &  {\color{blue} \cmark}      \\  
DADA-2000~\cite{fang2021dada}   &  1,962       &   649K       &      {\color{blue} \cmark}    &   \textbf{660}$\times$\textbf{1584}    &   \textbf{R}      &  {\color{blue} \cmark}   \\ 
GTACrash~\cite{kim2019crash, kim2021predicting} & 11,381 & 228K & - & 400$\times$710 & S &  {\color{blue} \cmark}  \\ 
YTCrash~\cite{kim2019crash, kim2021predicting} & 222 & 4,440 & -  & 400$\times$710& \textbf{R} & {\color{blue} \cmark}  \\ 
ROL~\cite{karim2023attention}   &  1,000       &  100K        &      - &   \textbf{720}$\times$\textbf{1080}         &    \textbf{R}     &   {\color{blue} \cmark}    \\  
TRA~\cite{liu2021temporal}   &   2,000      &    40-200K      &      -   & $\geq$\textbf{720}$\times$\textbf{1280}      &   \textbf{R}      &  A    \\  
DeepAccident~\cite{wang2024deepaccident}   &  -       &   57K       &    {\color{blue} \cmark}     &   \textbf{900}$\times$\textbf{1600}    &  S       & {\color{blue} \cmark}     \\  
DoTA~\cite{yao2020and, yao2022dota}   &    4,990     &    504K          & {\color{blue} \cmark}\tablefootnote{Re-annotated by MMAU~\cite{fang2024abductive}.} & \textbf{720}$\times$\textbf{1280}            &      \textbf{R}     & {\color{blue} \cmark}     \\ 
CTAD~\cite{luo2023simulation}   &   1,100     &   727K       &    {\color{blue} \cmark}    & 480$\times$640       &     S    &  {\color{blue} \cmark}   \\ \hline 
\textbf{Nexar} &  1,500       &  \textbf{1.70M}        &      {\color{blue} \cmark}  & \textbf{720}$\times$\textbf{1280}         & \textbf{R}       & {\color{blue} \cmark}   \\ \Xhline{4\arrayrulewidth}
\end{tabular}}
\label{tab:stats}
\end{table}

\begin{figure*}[h]
  \centering
   \includegraphics[width=0.98\linewidth]{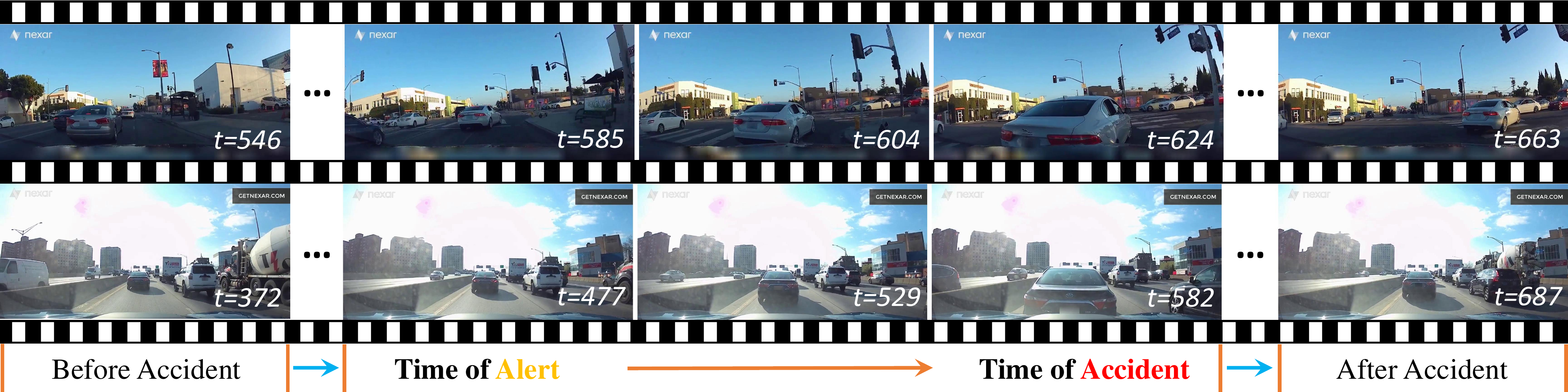}

   \caption{Nexar dataset samples. First column represents \textbf{Before accident interval}, columns 2-4 represent \textbf{Alert interval}, and fifth column represent \textbf{After accident interval}. Within the alert interval, column 2 represents the ''time-of-alert'' which is the earliest moment that the driver could intervene to prevent the accident. column 4 represents the ''time-of-accident'', and column 3 represents an intermediate frame between ``time-of-alert'' and ``time-of-accident''.}
   \label{fig:temporal}
\end{figure*}

A key challenge in accident prediction is its online nature. 
Unlike accident detection ~\cite{kamijo2000traffic}, where the entire video sequence is available offline, here the system needs to predict an event (accident) before it occurs. Poor judgment can result in failing to prevent or mitigate an accident (false negative), or taking unnecessary preventive actions (false positive) that may introduce new risks, such as sudden braking. Moreover, anticipation time is critical - detecting an accident too late, even if correct, may render the prediction ineffective. Therefore, both accuracy and anticipation time are crucial. This motivated us to propose an evaluation metric that integrates both aspects into a single score: the Average Precision over different times to the event.

Another challenge in accident anticipation is the subjectivity of determining when an accident can be predicted and when preventive action should be taken. This uncertainty complicates dataset construction, leading to multiple temporal labeling definitions, each reflecting different perspectives on when an accident begins~\cite{karim2023attention, fang2021dada, fang2024abductive, yao2019unsupervised, yao2020and, yao2022dota}. We define the ``time of alert'' as the earliest moment when a focused human can recognize that a dangerous situation is about to occur. To mitigate subjectivity, we derive the ``time of alert'' from annotations provided by multiple annotators. This ``time of alert'' serves as a baseline for setting expectations on when a computer-based model should predict the accident.

This paper presents the Nexar Dashcam Crash Prediction Challenge for evaluating early traffic accident anticipation, accompanied by a newly created dataset.  
Our temporal annotations for accidents and near-collisions divide videos into three intervals (Figure \ref{fig:temporal}):

1) \textbf{Before accident interval}, which starts at the beginning of the video and ends at the ``time of alert,'' the earliest moment a driver should intervene to prevent a possible accident.

2) \textbf{Alert interval}, which starts at the ``time of alert'' and ends at the ``time of accident/near accident,'' the actual collision time point and the last moment when anticipation is meaningful.

3) \textbf{After accident interval}, which begins after the collision and extends until the end of the video.

\begin{figure*}[h]
  \centering
   \includegraphics[width=0.98\linewidth]{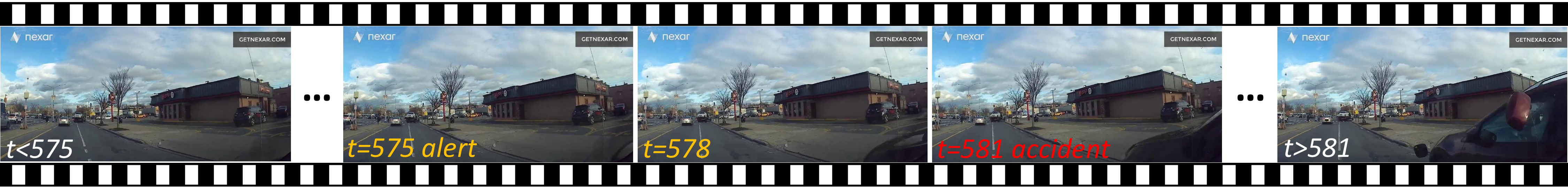}
    \caption{Our dataset contains a considerable fraction of samples where the accident happens within just a couple of frames, a time interval that is far shorter than a human driver could react. The anticipation for such cases is applicable only for autonomous driving scenarios, indicating that our dataset opens up the potential for evaluating algorithms that serve beyond human drivers.}
   \label{fig:pillar}
\end{figure*}

\begin{figure*}[h]
  \centering
   \includegraphics[width=0.98\linewidth]{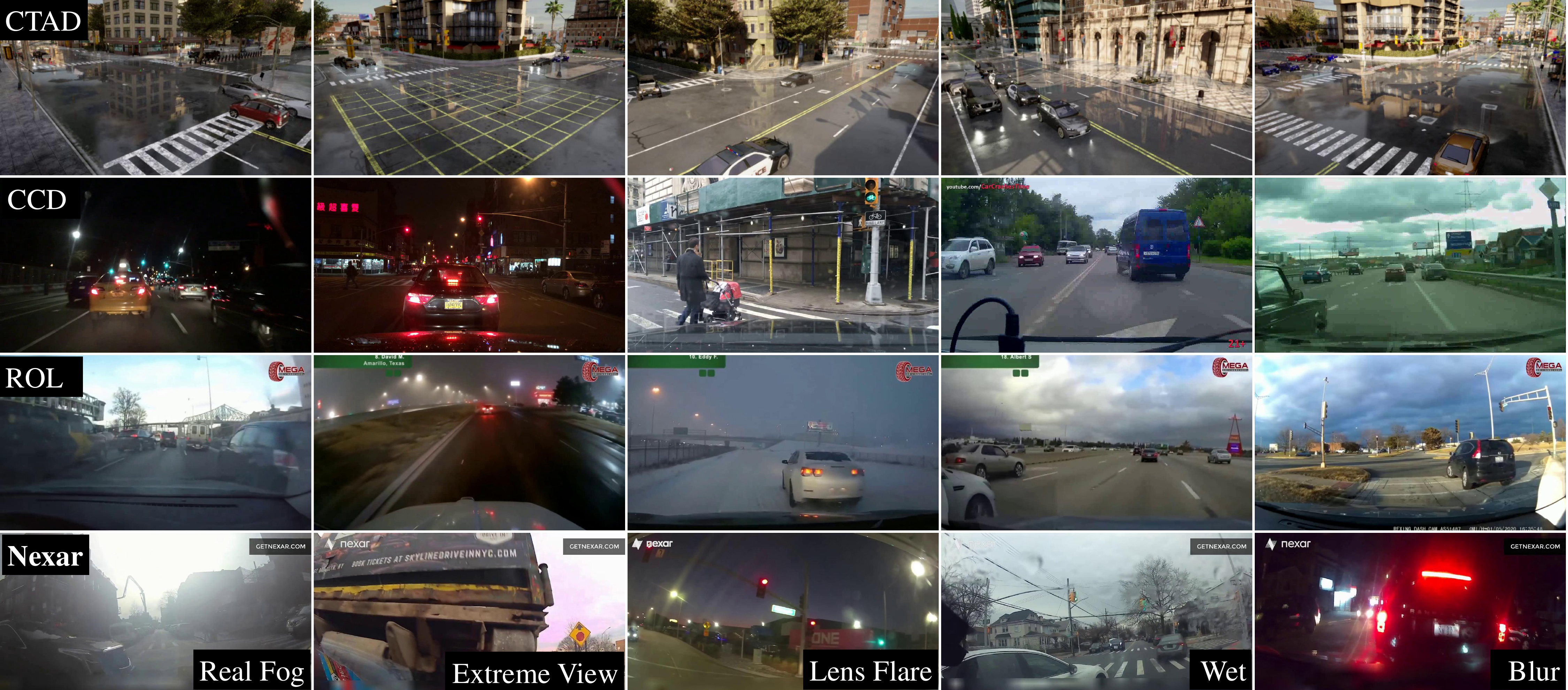}

   \caption{Our dataset demonstrates significant diversity regarding road types, weather condition, lighting conditions, types of vehicles as well as blurring, lens flare or other artifacts caused by the camera capture process, serving as a more challenging testbed than representative existing datasets for early traffic anticipation (CTAD~\cite{luo2023simulation}, CCD~\cite{bao2020uncertainty}, ROL~\cite{karim2023attention}).}  
   \label{fig:diversity}
\end{figure*}

Our dataset presents significant domain challenges compared to existing datasets. It includes a substantial portion of highly challenging videos where the alert interval lasts only a few frames, as illustrated in Figure \ref{fig:pillar}. Additionally, it captures various lighting conditions, weather situations, and road types, along with camera-related artifacts such as reflections, motion blur, fog, and lens flare (Figure \ref{fig:diversity}).

The rest of the paper is organized as follows. Section~\ref{sec:related} reviews the related literature. We introduce the dataset and its annotation process in Section~\ref{sec:dataset}. Section~\ref{sec:challenge} outlines the challenge details. We summarize our contributions in Section~\ref{sec:contributions} and conclude by suggesting potential future directions in Section~\ref{sec:future}.

\section{Related Work}
\label{sec:related}

\textbf{Traffic Accident Anticipation}. 
The challenge presented in this paper concerns Traffic Accident Anticipation (TAA)~\cite{hu2003traffic, zeng2017agent, chan2017anticipating, suzuki2018anticipating, corcoran2019traffic, fatima2021global, bao2020uncertainty, karim2022dynamic, bao2021drive, kim2021predicting, malawade2022spatiotemporal, karim2022toward, karim2023attention, fang2022cognitive, li2024cognitive, wang2023gsc, mahmood2023new, liu2023learning, maruyama2023accident, liu2023net, song2024dynamic, jeongdriver, li2023traffic, al2024traffic, verma2024vision, liao2024real, liao2024and, liao2024crash, liu2025ccaf, patera2024spatio, qiao2024msan, harada2025traffic}. TAA differs from Traffic Accident Detection (TAD)~\cite{kamijo2000traffic} in that TAA requires early anticipation of an accident before observing future sequences, making it significantly more challenging than determining when an accident happens in a complete video sequence.
Several efforts have been made to address the TAA task.
Early works, including \cite{hu2003traffic}, prior to the deep learning era, anticipated potential traffic accidents using vehicle trajectory cues.
Chan et al.~\cite{chan2017anticipating} formally defined the task, introduced an evaluation metric, and proposed the first large-scale dataset for evaluation.
This was followed by Zeng et al.~\cite{zeng2017agent} and Suzuki et al.~\cite{suzuki2018anticipating}, who introduced novel architectural designs and loss functions.
Most approaches~\cite{bao2020uncertainty, malawade2022spatiotemporal, chan2017anticipating, zeng2017agent, suzuki2018anticipating, kim2021predicting} follow a framework where a feature extraction network captures essential spatial and temporal cues, followed by a recurrent unit that accumulates and propagates temporal information to generate predictions.
Other approaches leverage region-based feature extraction followed by a graph convolutional network to model inter-relations, as proposed in \cite{bao2020uncertainty, malawade2022spatiotemporal}.
Bao et al.~\cite{bao2021drive} also applied a reinforcement learning scheme to predict potential accident risks.
With the rise of vision transformer architectures~\cite{vaswani2017attention}, which have proven effective in various vision tasks~\cite{liu2021swin, wang2025internvideo2, dosovitskiy2020image, touvron2021training, carion2020end}, self-attention mechanisms have also been employed for accident anticipation~\cite{karim2022dynamic, karim2023attention, song2024dynamic}.

\textbf{Dashcam-based datasets for traffic accident anticipation}.  
To facilitate research and evaluation, various datasets have been introduced alongside the development of algorithms for TAA~\cite{baee2021medirl, herzig2019spatio, chan2017anticipating, zeng2017agent, aliakbarian2018viena, kataoka2018drive, yao2019unsupervised, you2020traffic, bao2020uncertainty, fang2021dada, kim2019crash, karim2023attention, liu2021temporal, wang2024deepaccident, yao2022dota, fang2024abductive}.  
In general, datasets for TAA require both positive (accident occurred) and negative (no accident) video clips, in contrast to the positive-only nature of temporal localization tasks in TAD.  
Various datasets provide additional annotations, including spatial annotations~\cite{bao2020uncertainty}, temporal annotations~\cite{fang2021dada, fang2024abductive}, scene attribute annotations~\cite{bao2020uncertainty, fang2021dada, luo2023simulation, fang2024abductive}, driver attention labels~\cite{fang2021dada}, accident categories~\cite{fang2022cognitive}, as well as language-based descriptions and reasoning~\cite{fang2024abductive}.  
Synthetic datasets~\cite{aliakbarian2018viena, wang2024deepaccident, luo2023simulation, kim2019crash} have also been introduced, offering advantages such as large-scale data availability and fully annotated ground truth.  
Table \ref{tab:stats} provides statistics and additional properties of these datasets.  

Comparing to existing datasets, our dataset provides significantly more frames per video, offering a dual advantage: a richer temporal context for accident anticipation and a more comprehensive feature space, enabling models to better distinguish between predictive cues that reliably indicate accidents and those that do not necessarily lead to a collision.

\textbf{Temporal Annotations for Accident Anticipation}.  
Existing datasets explore different temporal annotations of accidents. Prior studies can be categorized into two main approaches based on their criteria for defining the start and end times of an event, as summarized in Table \ref{tab:temporal}.  

\text{1)} \textbf{Appearance-based}~\cite{karim2023attention, fang2021dada, fang2024abductive}, where the start time is defined as the initial full or partial appearance of the vehicle that eventually causes the accident ($t_\text{appearing}$), and the end time is set as the collision moment ($t_\text{collision}$).  

\text{2)} \textbf{Anomaly- or causality-based}~\cite{yao2019unsupervised, yao2020and, yao2022dota, you2020traffic}, where the start time corresponds to the moment the annotator perceives the accident as inevitable or when the vehicle begins to exhibit abnormal behavior ($t_\text{inevitable}$), while the end time is defined as the point when all vehicles are either out of sight or stationary ($t_\text{end\_of\_anomaly}$).  

While \text{1)} provides a more objective annotation, it is biased toward earlier timestamps since the mere appearance of a third-party vehicle does not necessarily indicate the moment when the accident becomes predictable.  
Conversely, \text{2)} is subjective and often lacks precise temporal annotation for the exact collision moment.  

Although MMAU~\cite{fang2024abductive} has made significant efforts to re-annotate several datasets and bridge the gap between different temporal annotation strategies, its annotations are only available for a subset of samples within each dataset and may still introduce bias when combined with the original temporal annotations.  

We believe our annotation approach offers the best combination of the advantages of both ``anomaly-based'' and ``appearance-based'' temporal definitions. Specifically, $t_\text{inevitable}$ provides a reasonable starting point for accident prediction, while $t_\text{collision}$ marks the last meaningful moment for anticipation.  
To address the subjective nature of appearance-based alert time annotation, we calculate statistics over annotations from multiple annotators.

\begin{figure}[h]
  \centering
   \includegraphics[width=0.98\linewidth]{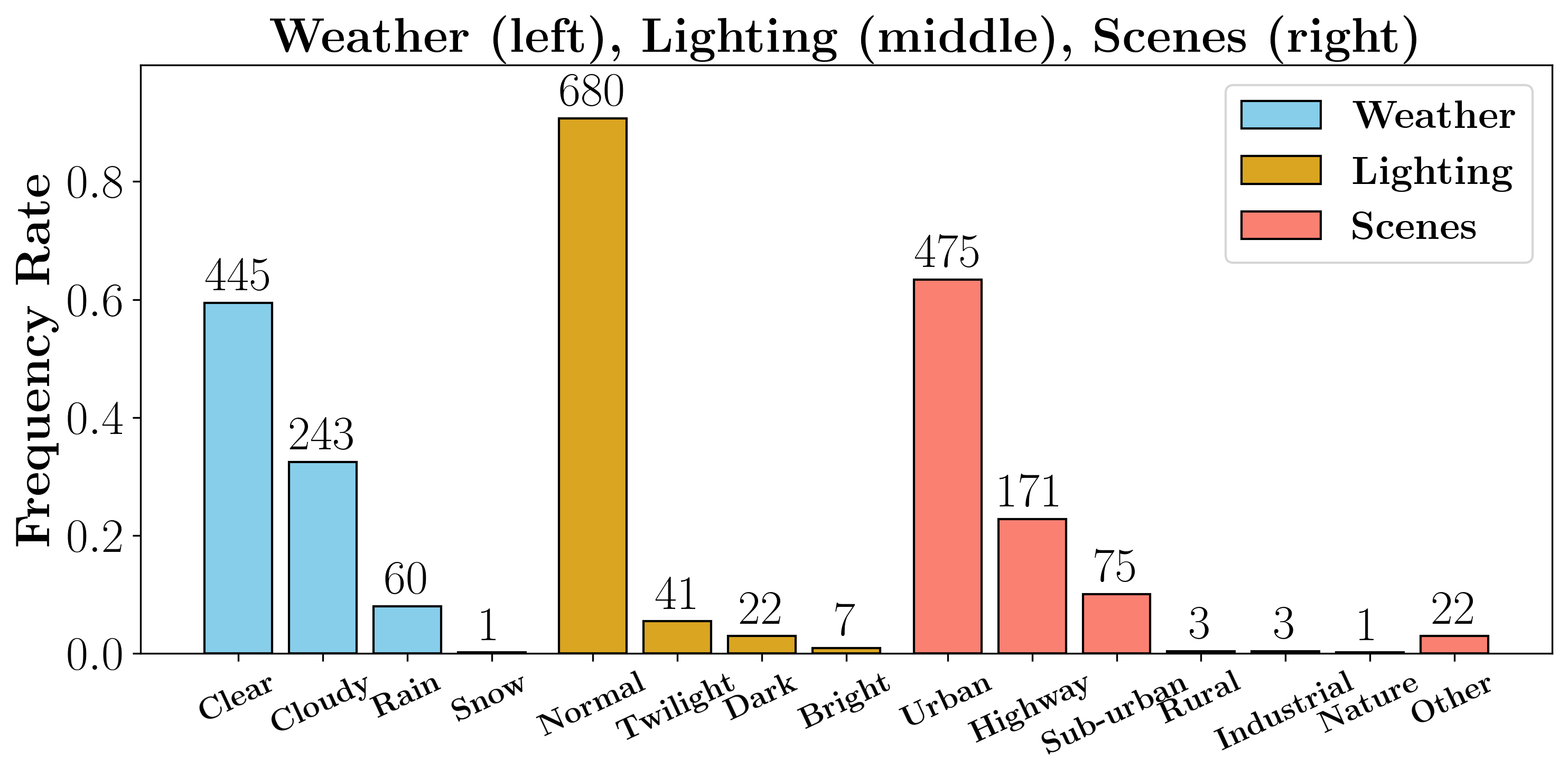}

   \caption{Our dataset demonstrates diversity regarding video recording with respect to the weather, lighting condition as well as road types. One of the video was recorded indoor and does not have the weather label.}
   \label{fig:attributes}
\end{figure}

\begin{table}[htbp]
\caption{Different datasets provide different definition of the temporal annotations. ``{\color{black} \cmark}'' represents available temproal annotations from the original paper release, while ``{\color{black} M}'' represents re-annotation by MMAU~\cite{fang2024abductive} (only part of the samples were re-annotated).}
\resizebox{0.48\textwidth}{!}{\begin{tabular}{lcccc}
\Xhline{4\arrayrulewidth}
      & $t_\text{appearing}$ & $t_\text{inevitable}$  & $t_\text{collision}$ & $t_\text{end\_of\_anomaly}$\\ \hline
A3D~\cite{yao2019unsupervised} & {\color{black} M} & {\color{black} \cmark} & {\color{black} M} & {\color{black} \cmark}, {\color{black} M}\\ 
DoTA~\cite{yao2020and, yao2022dota} & {\color{black} M} & {\color{black} \cmark} & {\color{black} M} & {\color{black} \cmark}, {\color{black} M}\\ 
CTA~\cite{you2020traffic} & - & {\color{black} \cmark}\tablefootnote{This is based on the definition of the starting of the ``causality'', the moment when the vehicle that caused the eventual accident starts to demonstrate wrongdoing and eccentric behavior. We observe that such definition is slightly late-biased compared to the definition of $t_\text{inevitable}$. } & {\color{black} \cmark} & - \\
CCD~\cite{bao2020uncertainty} & {\color{black} M} & - & {\color{black} \cmark}, {\color{black} M} & {\color{black} M}\\
ROL~\cite{karim2023attention} & {\color{black} \cmark} & - & {\color{black} \cmark} & - \\
DADA2000~\cite{fang2022cognitive} & {\color{black} \cmark} & - & {\color{black} \cmark} & {\color{black} \cmark} \\
\hline 
\textbf{Nexar} & - & {\color{black} \cmark}\tablefootnote{Note our formal definition is the earliest moment a driver should intervene to prevent the accident, which is slightly different from the original definition of $t_\text{inevitable}$.} & {\color{black} \cmark} & - \\ \Xhline{4\arrayrulewidth}
\end{tabular}}
\label{tab:temporal}
\end{table}

\section{Dataset Description}
\label{sec:dataset}


Our dataset consists of 1,500 road facing videos recorded by Nexar dashcams. Video recording is triggered by hard-breaks or sudden accelerations detected by the in-camera IMU. Each video has a resolution of 1280×720 at approximately 30 frames per second and an average duration of 40 seconds.

A team of annotators reviewed hundreds of thousands of videos and labeled them based on the event that triggered the recording as:
\begin{itemize}
    \item Collision: when there is an accident event between the ego vehicle and a third party;
    \item Near-collision: when there is a close call event between the ego vehicle and a third party;
    \item Normal driving: when there is no collision or near-collision (no event).
\end{itemize}

Collisions and near-collisions are considered critical events and are grouped together as positive examples of situations that can lead to an accident, while normal driving videos serve as negative examples.  

Two annotation tasks were conducted:  
1) A \textbf{general attributes annotation task}, in which all videos were classified based on weather, scene, and lighting conditions.  
2) A \textbf{incidents specific annotation task} for positive examples, which involved identifying whether the event was visible in the video, determining the event time, and annotating the alert time.

Categorical attributes were annotated by three annotators, with the final value determined as the mode of the annotations.

Temporal attributes were annotated by 10 different annotators. The final event time was determined as the median value of all annotations, while the final alert time was set as the second-highest value. Given the high subjectivity of alert time annotations, we selected the second-highest value instead of the median to promote shorter alert-to-accident intervals and achieve higher consensus among annotators. The second-highest value also provides greater robustness to outliers compared to the maximum.

\subsection{Acceptance criteria}

The dataset was filtered to include only videos that met the following criteria:
\begin{enumerate}
    \item At least 2 annotators agreed on all categorical attributes (e.g. event visibility, weather, scene, lighting conditions);
    \item The video was not corrupted;
    \item For positive cases, we required that the event be visible in the video, involve cars or trucks (excluding pedestrians, bicycles, motorcycles, animals, or stationary objects), and that the (potential) impact occur on the front side of the ego vehicle.
\end{enumerate}


\subsection{Video anonymization and data privacy}
While preparing this dataset, we prioritized the protection of privacy and compliance with ethical standards.
To ensure the privacy of drivers and other individuals captured in the videos, we applied the following anonymization measures:
\begin{enumerate}
    \item Blurring of faces, license plates, and the dashboard of the ego vehicle;
    \item Removal of audio from all videos;
    \item Exclusion of videos that begin or end within 100 meters or 2 minutes of the start or end of a Nexar user’s ride to prevent potential disclosure of sensitive locations such as home or workplace.
\end{enumerate}



These measures ensure that the dataset aligns with ethical
guidelines for research while minimizing the risk of re-identification or misuse. By openly addressing these considerations, we aim to encourage responsible use of this
dataset within the research community.

\subsection{Dataset sampling and split}

At the end of the filtering process, we randomly sampled 750 negative examples and 750 positive examples (400 collisions and 350 near-collisions), resulting in a balanced dataset with 1,500 videos. In addition, we reserved a part of the videos to make a test set (more details in section \ref{ss:test_set}).
\subsection{Dataset statistics}

\begin{figure*}[h]
  \centering
   \includegraphics[width=0.98\linewidth]{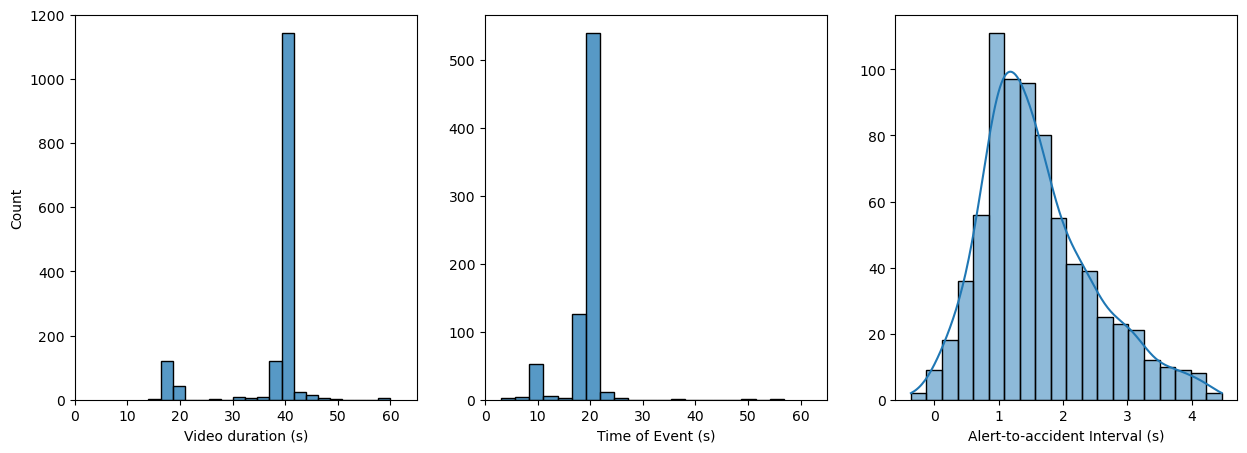}

   \caption{Histogram of video duration (left), time of event (center), and alert-to-accident interval (right).}
   \label{fig:time_stats}
\end{figure*}

We illustrate the statistical distribution of the dataset in Figure \ref{fig:time_stats}. The distribution of video durations is bimodal, with most videos lasting approximately 40 seconds. The duration depends on the dashcam model, as some models record shorter clips. Events (collisions and near-collisions) typically occur near the middle of the video. The average alert-to-accident interval is 1.6 seconds, with a maximum of 4.5 seconds.  

Regarding scene, weather, and lighting conditions, our dataset reflects the distribution captured by our cameras, with urban scenes, clear skies, and normal lighting conditions being the most frequent.

The dataset is available on huggingface \footnote{https://huggingface.co/datasets/nexar-ai/nexar\_collision\_prediction} and provided subject to the Nexar license \footnote{https://huggingface.co/datasets/nexar-ai/nexar\_collision\_prediction/blob/main/LICENSE}


%
%


\section{Challenge Description}
\label{sec:challenge}

The challenge aims to advance the development of accident anticipation models using real-world dashcam footage. Participants should build machine learning models to estimate the likelihood of an imminent collision based on short video segments. Hosted on Kaggle~\cite{nexar2025challenge}, the competition consists of a single task: predicting whether an accident is about to occur within a given time frame.

\subsection{Test set}
\label{ss:test_set}
The competition test set is composed of 1344 videos with a duration of approximately 10 seconds. These videos were generated from a pool of 568 videos that follow a distribution similar to that of the training data. For each positive video, up to 3 videos are generated by cropping the original video so that videos have approximately 10 seconds and terminate at instants $t_{\text{event}} - \Delta_{\text{TTE}}$, where $t_{\text{event}}$ is the time of the (near-)collision event in milliseconds and $\Delta_{\text{TTE}}$ represents different times to event, namely 500ms, 1000ms and 1500ms. This allows testing how models behave at different times to accident. A video is only cropped at a given $\Delta_{\text{TTE}} > 500$ when the difference between the annotated event and the alert time is greater than the correspondent $\Delta_{\text{TTE}}$. For negative examples, fake event times were generated by adding Gaussian noise to half of the duration of the video, and by generating fake alert times to match the distribution of the positive cases.

The test set is subdivided into two balanced sets of equal size: public and private. The public test set is used to evaluate model submissions during the competition. The private test set will be used when the competition closes to make the final leader board. Teams continuously submit solutions to the entire test set without knowing the internal division between public and private sets. 

\subsection{Evaluation metric}
The goal of the competition is to find solutions that maximize recall, precision, and anticipation time (i.e. detect positive cases as early as possible). We use the mean of the Average Precision (AP) measured at different time-to-event (TTE) values. The AP summarizes the relation between Precision and Recall at a fixed TTE. By combining the AP measured at different TTE values we can summarize the models' ability to anticipate a potential collision event.

The output of the model should be a confidence score for each of the videos in the test set. Scores are grouped into three sets according to the time-to-event (TTE)  of the video (500 ms, 1000 ms, or 1500 ms). Then, for each group, the Average Precision (AP) is computed. AP measures the area under the precision-recall curve as the weighted mean of precisions achieved at each threshold, with the increase in recall from the previous threshold used as the weight (scikit-learn implementation~\cite{pedregosa2011scikitlearn}):
\begin{equation}
\text{AP} = \sum_n (R_n - R_{n-1}) P_n\text{ ,}
\end{equation}
where $P_n$ and $R_n$ are the precision and recall at the nth threshold. Having the AP calculated for each TTE, the mean Average Precision is calculated as the mean of the three AP scores:
\begin{equation}
\text{mAP} = \frac{\text{AP}_{500}+\text{AP}_{1000}+\text{AP}_{1500}}{3}\text{ .}
\end{equation}

\section{Contributions}
\label{sec:contributions}

In this paper, we introduced the Nexar Dashcam Collision Prediction Dataset and Challenge, aimed at advancing research in traffic accident anticipation and autonomous driving safety. 

Our key contributions are as follows.
\begin{itemize}
    \item We release to the community, under a permissive license, a new challenging and diverse dataset for training and evaluating  algorithms for traffic accident anticipation. The dataset is captured from multiple urban and suburban areas in the US and encompasses a wide range of real-world scenarios, including various weather and lighting conditions, scene types, and dashcam capturing effects. Our clips consist of high-definition (720p) video sequences, each spanning approximately 40 seconds at 30 frames per second, providing rich temporal context that enables models to identify reliable accident predictors.
    \item We provide consistent and well-structured temporal annotations tailored for the traffic accident anticipation task. Our dataset includes a significant portion of cases where the accident occurs within just a few frames -- far shorter than a human driver's reaction time -- making it particularly valuable for evaluating algorithms in the context of autonomous driving. To address the subjectivity of alert time annotation, we use multiple annotators and extract robust statistics. 
    \item The challenge establishes a standardized evaluation framework and benchmarking protocol available in Kaggle~\cite{nexar2025challenge}, emphasizing early and reliable predictions through multiple precision-recall curves across different time-to-event intervals. 

\end{itemize}


\section{Future Directions}
\label{sec:future}

We believe this work makes a significant contribution to traffic safety and autonomous vehicle research by fostering innovation in early accident prediction. Future improvements to the dataset could include expanding annotations to capture additional contextual factors such as vehicle types, driver behaviors, and road infrastructure details. Increasing the dataset’s diversity by incorporating footage from different geographic regions and traffic environments would further enhance its robustness. Additionally, integrating edge-case scenarios, such as extreme weather conditions, events involving vulnerable road users, and complex multi-vehicle interactions, would make the dataset more comprehensive for real-world applications.

Beyond dataset expansion, this dataset can serve as a benchmark for future automated accident prediction systems, particularly for autonomous vehicles and ADAS technologies. The dataset includes a significant number of cases where the alert-to-accident interval is extremely short, making it especially valuable for testing accident prevention in real-time decision-making scenarios. Given that human reaction time is typically at least 0.7 seconds ~\cite{green2000long}, many of these accidents would be unavoidable for human drivers. However, an autonomous vehicle equipped with real-time accident prediction capabilities could anticipate and intervene earlier. Moreover, in a V2X-enabled environment, the detecting vehicle could alert surrounding vehicles, enabling a coordinated response to mitigate or even prevent collisions. This positions our dataset as a foundation for advancing accident prediction and prevention, accelerating progress toward safer, more proactive transportation systems.

{\small
\bibliographystyle{ieee_fullname}
\bibliography{egbib}
}

\end{document}